\documentclass{article}
\usepackage{iclr2027_conference,times}
\iclrfinalcopy
\usepackage{amsmath,amssymb,mathtools,booktabs,graphicx,placeins,multirow,threeparttable}
\usepackage[font=small,labelfont=bf]{caption}
\usepackage[colorlinks=true,linkcolor=blue,citecolor=blue,urlcolor=blue]{hyperref}
\usepackage{xspace}

\title{PQ-HSA: Reusing Product-Quantized Scores\\for Hybrid Sparse-Approximate Attention}

\author{Kunming Shao$^{1,*}$, Jierun Chen$^{2}$, Yanli Wang$^{3}$, Ruoyu Wang$^{4}$,\\
\textbf{Haoli Bai$^{2}$, Kwang-Ting Cheng$^{1}$, Chi Ying Tsui$^{1}$}\\[4pt]
{\normalfont $^{1}$The Hong Kong University of Science and Technology\quad
$^{2}$Huawei Technologies Ltd.}\\
{\normalfont $^{3}$Sun Yat-sen University\quad
$^{4}$Nanyang Technological University}\\[2pt]
{\normalfont $^{*}$Corresponding author: \texttt{kshaoaa@connect.ust.hk}}}

\begin{document}
\maketitle

\begin{abstract}
At each decoding step a language model attends over the key--value (KV)
cache of every earlier token, so at long context the attention call is
bounded by memory bandwidth. Sparse attention reads only a subset of keys
chosen by a cheap score estimate, and most methods give the unread tokens zero
weight. The output then draws on only a small fraction of the KV cache, and
accuracy drops at small budgets, most on tasks that aggregate information
across the context. An inverted-file product-quantization (IVF-PQ) index over the cached
keys computes an approximate score for every indexed token in order to rank
them; after ranking, those scores approximate the attention logits of the
tokens left out. PQ-HSA (hybrid sparse-approximate
attention) attends the selected tokens with their original keys and values,
and the unselected tokens, the background, enter the same softmax through
those scores, summed per inverted list and multiplied by the list's mean
value. At 128K and a 1--2\% retrieval budget, PQ-HSA is more accurate
than Quest and SnapKV on Llama-3.1-8B and Qwen3-30B-A3B and stays close to full attention in macro accuracy; with the same selector, the background term raises macro
accuracy on the 8B model from 0.71 to 0.83. In the same 128K setting, inside vLLM on one NVIDIA H20, the decode
attention call runs $1.6\times$ faster than the FlashAttention-3 kernel; the
speedup grows with context length, and a cost model fitted on 8B to 30B
models gives the context length at which it begins. A vLLM plugin runs PQ-HSA on two engine versions without
changes to the engine source; code is available at \url{https://github.com/KunmingSHAO/pqhsa_release}.
\end{abstract}

\section{Introduction}
\label{sec:introduction}

A decoder-only language model generates one token at a time, and at each
step the new query attends over the keys and values of every earlier token,
stored in the key--value (KV) cache \citep{dao2022flashattention}. The cache
grows linearly with the context, so at long context the attention call is
bounded by the memory bandwidth needed to read it. Yet the softmax of a
single query is often concentrated on a small fraction of the tokens
\citep{zhang2023h2o,tang2024quest}. Sparse attention builds on this: it
attends only to a subset of the cached tokens, chosen by an inexpensive
estimate of their attention scores, and so reads a small part of the cache
at each step.

Quest bounds the logits of each page of keys and keeps the highest-bound
pages \citep{tang2024quest}; SnapKV keeps the tokens that the end of the
prompt attended to most during prefill \citep{li2024snapkv}; PQCache,
RetrievalAttention and ParisKV build an approximate nearest-neighbor (ANN)
index over the keys and retrieve the top-$k$ per query
\citep{zhang2024pqcache,liu2024retrievalattention,qi2026pariskv}; MagicPIG
samples tokens with locality-sensitive hashing \citep{chen2024magicpig}.
In the selection-based methods the softmax is normalized over the selected
tokens only and the unselected tokens receive zero weight, which we refer to
as truncation. When attention is spread over many tokens, as in tasks that
aggregate evidence across the context, the omitted mass can carry part of
the answer; among the methods above, MagicPIG accounts for it through the
importance weights of its sample.

An IVF-PQ index \citep{jegou2011pq,johnson2017faiss} groups the keys into
$n_{\mathrm{list}}$ clusters by $k$-means, the inverted lists (the inverted
file, IVF), and compresses each key's offset from its cluster centroid into
$M$ short codes (product quantization, PQ); a query's inner product with any
indexed key is then approximated by $M$ table lookups.

\begin{figure}[t]
\centering
\includegraphics[width=\linewidth]{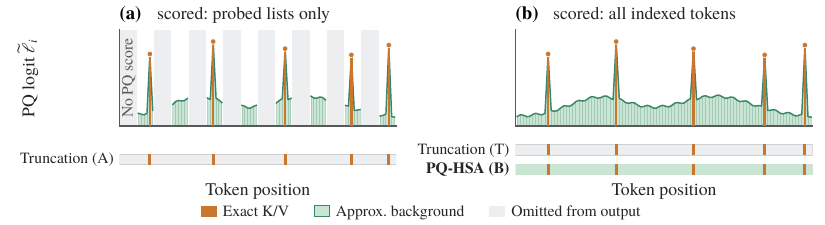}
\caption{Token coverage (schematic). (a) Only the probed IVF lists are scored, as in standard ANN retrieval, and
truncation (variant A) attends the selected tokens alone. (b) Every indexed
token is scored; variant T attends only the top-$k$ tokens, and PQ-HSA
(variant B) adds the other scored tokens as an approximate background. The strips show contribution to
the output, not exact KV reads.}
\label{fig:coverage}
\end{figure}

\begin{figure}[t]
\centering
\includegraphics[width=\linewidth]{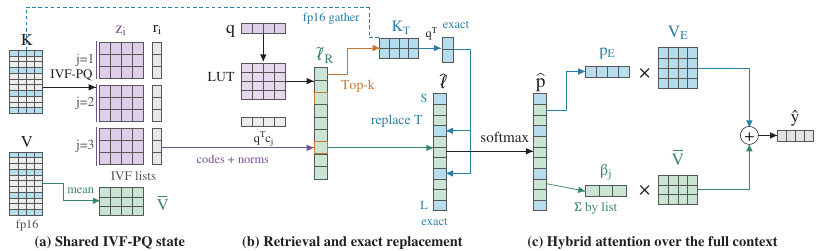}
\caption{PQ-HSA shares one IVF-PQ score stream between retrieval and
background attention. (a) Residual PQ codes, key norms and list value means
are kept alongside the resident original KV cache. (b) The scan scores every
indexed token in $R$ from a lookup table (LUT) built from the query; the
top-$k$ tokens $T$ are gathered and their logits replaced by exact ones.
(c) The exact set $E=S\cup L\cup T$ and the background $B=R\setminus T$
share one softmax: selected weights multiply original values, and background
weights are summed per list and multiply list means. Tensor sizes are
schematic.}
\label{fig:framework}
\end{figure}

PQ-HSA (hybrid sparse-approximate attention) scans the codes of every indexed key at each
decode step instead of probing a few lists, so one scan gives an approximate
attention logit for every token, and PQ-HSA uses these logits twice. The
top-scoring tokens are selected and attended with their original keys and
values. Every other token, the background, keeps its approximate logit in
the same softmax; these weights are summed per inverted list and multiplied by the list's mean
value. The softmax denominator and the output thus account for the whole
context, while only the selected KV rows are read in full
(Figures~\ref{fig:coverage} and~\ref{fig:framework}). No second scoring pass
is needed; the background adds one sum per list and a read of
$n_{\mathrm{list}}$ value means (Section~\ref{sec:method-operator}).

The construction is not specific to IVF-PQ: any index that scores every
cached key cheaply yields both a selection and an estimate for the keys it
leaves out. Because IVF-PQ is a classic vector-search index, the many
optimizations developed for it can be brought into attention directly
\citep{ge2014optimized,guo2020accelerating,douze2026faiss}.

Three observations organize the evaluation. First, at the same budget
PQ-HSA is more accurate than Quest, SnapKV and ANN retrieval baselines on
Llama-3.1-8B and Qwen3-30B-A3B at 128K, and stays close to full attention
(Section~\ref{sec:quality}). Second, with the same selector and budget,
the background term raises macro (mean per-task) accuracy by 0.12 on
Llama-3.1-8B and also raises it on Qwen3-30B; the
background needs both the PQ logits and the list value means
(Section~\ref{sec:mechanism}). Third, in the same 128K setting inside vLLM the
attention call is $1.6\times$ faster than the dense FlashAttention-3 kernel
\citep{shah2024flashattention3}, rising to $2.7\times$ at 512K on a
1M-window checkpoint. A cost model fitted on 8B to 30B models gives each
model's break-even length from its number of KV heads and its group size
under grouped-query attention (GQA), where a group of query heads shares one
KV head \citep{ainslie2023gqa} (Section~\ref{sec:speed}).

PQ-HSA builds on published components: IVF-PQ search
\citep{jegou2011pq,johnson2017faiss} with the key norm kept outside the
quantizer \citep{dai2019neq}, attention sinks (the first tokens) with a local
window \citep{xiao2023streamingllm}, ANN retrieval of KV rows at decode time
\citep{zhang2024pqcache,liu2024retrievalattention,qi2026pariskv}, batched
$k$-means \citep{flashkmeans2025} and vLLM's paged KV storage in fixed-size
blocks \citep{kwon2023vllm}. Our contributions are:
(1)~a construction of hybrid sparse--approximate attention in which the
scores a vector-search index computes for ranking also serve as the logits of
the unselected tokens; with IVF-PQ, they are aggregated per list with list
value means under one normalizer (Section~\ref{sec:method});
(2)~a controlled mechanism study that isolates the background term and the
roles of its logits and list value means under the same selector and budget,
on two model families
(Section~\ref{sec:mechanism});
(3)~an empirical cost model for decode sparse attention across model sizes
and GQA group sizes, with fitted break-even lengths (Section~\ref{sec:speed});
(4)~a vLLM plugin that leaves the engine source unchanged, reads the paged
KV cache in place, and runs on vLLM 0.8.5 and 0.29
(Section~\ref{sec:deployment}); the operator, kernels and plugin are
released as open source.

\begin{figure}[t]
\centering
\includegraphics[width=\linewidth]{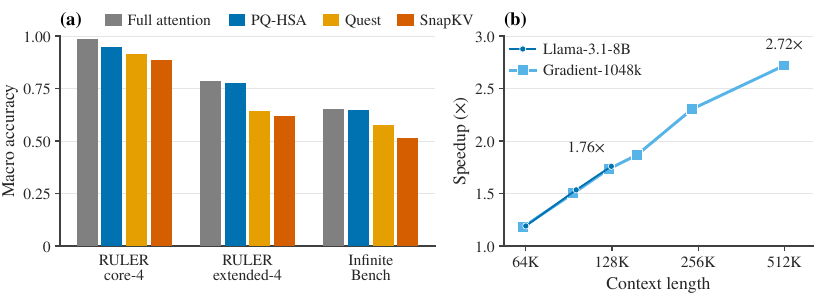}
\caption{(a) Macro accuracy of Llama-3.1-8B-Instruct at a 1\% retrieval
budget on the RULER core-4 and extended-4 task groups (Section~\ref{sec:setup}) and InfiniteBench at 128K. (b) Speedup of the decode attention
call (summed over layers) over the FlashAttention-3 decode kernel
\citep{shah2024flashattention3} on one H20.
Gradient-1048k, a Llama-3-8B checkpoint with a 1M-token window, has the
architecture of Llama-3.1-8B (8 KV heads, $G=4$) and extends the sweep
beyond Llama-3.1's 128K window.}
\label{fig:headline}
\end{figure}

\section{Related Work}
\label{sec:related}

Decode-time methods that reduce the KV read can be grouped by what happens to
the tokens they do not read.

\paragraph{Evicting.}
StreamingLLM keeps the first tokens, which act as attention sinks, and a
recent window \citep{xiao2023streamingllm}; H2O and TOVA drop tokens by
accumulated or recent attention \citep{zhang2023h2o,oren2024tova}; SnapKV
selects once at the end of prefill from the attention of an observation
window \citep{li2024snapkv}. A dropped token is unavailable to later
queries.

\paragraph{Selecting.}
Quest scores each page of keys by an upper bound on its logits and attends
to the top pages \citep{tang2024quest}; PQCache, RetrievalAttention and
ParisKV index the keys with ANN structures and retrieve the top-$k$ per
query, ParisKV with a low-bit second-stage rerank
\citep{zhang2024pqcache,liu2024retrievalattention,qi2026pariskv}. Loki,
Double Sparsity and HashAttention rank tokens from low-rank keys,
offline-calibrated channels and learned binary signatures
\citep{singhania2024loki,yang2024doublesparsity,desai2025hashattention};
Squeezed Attention and ClusterKV cluster the keys and keep whole clusters by
centroid score \citep{hooper2025squeezed,liu2025clusterkv}; InfLLM looks up
blocks of past tokens through representative keys \citep{xiao2024infllm};
InfiniGen, ShadowKV and ClusterKV keep the KV cache, or its values, in host
memory and fetch the selected rows
\citep{lee2024infinigen,sun2025shadowkv,liu2025clusterkv}. Speculative
token selection runs a draft model at every step to nominate the tokens
\citep{sts2026}. Each of these normalizes the softmax over the selected
tokens, so the unselected tokens receive zero weight.

\paragraph{Approximating the remainder.}
Splitting attention into an exact sparse part and an approximation of the
rest goes back to Scatterbrain, which adds a low-rank estimate, and to
clustered attention, which groups the queries; both approximate
full-sequence attention rather than decoding over a KV cache
\citep{chen2021scatterbrain,vyas2020clustered}. At decode time, SparQ scores
every token from a few query-selected key components and gives the mass
outside its top-$k$ to one global mean value \citep{ribar2024sparq}.
RetroInfer and Multipole Attention cluster the keys, attend the
highest-scoring clusters exactly and represent the other clusters by a
centroid logit and an aggregated value, so both selection and approximation
work at cluster granularity; RetroInfer also keeps the KV cache in host
memory and drops the lowest-ranked clusters
\citep{chen2026retroinfer,hooper2025multipole}. MagicPIG samples tokens with
locality-sensitive hashing and forms an importance-weighted estimate,
computed on the CPU \citep{chen2024magicpig}.

\paragraph{Quantizing.}
Product quantization encodes vectors as short codes in subspaces
\citep{jegou2011pq,johnson2017faiss}, norm-explicit quantization keeps the
norm outside the quantizer \citep{dai2019neq}, and Transformer-VQ, CommVQ
and self-indexing caches store quantized keys
\citep{lingle2024transformervq,li2025commvq,yang2026selfindexing}.

PQ-HSA is a selecting method whose unselected tokens keep the
query-dependent PQ scores the selector already computed and contribute
through their list's value mean, under the same normalizer as the selected
set. Unlike SparQ it uses one value mean per list rather than one global
mean, and unlike the clustered methods it selects individual tokens from a scan of
every key, with the KV cache kept on the GPU. It shares the IVF-PQ index with PQCache, which normalizes over the retrieved
tokens alone; Appendix~\ref{app:design} compares these designs side by side.

\section{The Proposed Product-Quantization based Hybrid Sparse-Approximate Attention (PQ-HSA)}
\label{sec:method}

For a decode query $q\in\mathbb{R}^d$ and cached keys and values
$\{(k_i,v_i)\}_{i=1}^{n}$, dense attention computes
$y=\sum_i e_i v_i/\sum_i e_i$ with $e_i=\exp(q^\top k_i/\sqrt d)$. For any
split of the positions into an exact set $E$ and its complement $B$, the
standard decomposition
\begin{equation}
 y=\frac{N_E+N_B}{Z_E+Z_B},\qquad
 N_X=\sum_{i\in X}e_iv_i,\quad Z_X=\sum_{i\in X}e_i
 \label{eq:decomp}
\end{equation}
holds. A sparse method chooses $E$ and decides what to do with $N_B$ and
$Z_B$: truncation sets both to zero; PQ-HSA estimates both from the
approximate scores that chose $E$.

\subsection{Direction--norm residual IVF-PQ over keys}
\label{sec:method-index}

The index is standard IVF-PQ \citep{jegou2011pq,johnson2017faiss} applied to
key directions. Each key is written $k_i=r_iu_i$ with norm $r_i$ and unit
direction $u_i$. The directions are
clustered into $n_{\mathrm{list}}$ inverted lists with centroids $c_j$, and
each residual $u_i-c_{a_i}$ is product-quantized in $M$ subspaces with $2^b$
codewords $p_{m,z}$ each, so that a key is stored as its list id $a_i$, its
codes $z_{i1},\dots,z_{iM}$ and its norm. For a query $q=r_qu_q$ the index
returns, for every indexed position,
\begin{equation}
 \widetilde\ell_i(q)=\frac{r_qr_i}{\sqrt d}\Big(u_q^\top c_{a_i}
 +\sum_{m=1}^{M}\big(u_q^{(m)}\big)^{\!\top}p_{m,z_{im}}\Big),
 \label{eq:pqscore}
\end{equation}
computed as one table lookup per subspace once an $M\times 2^b$ lookup table
(LUT) of the inner products of each subvector $u_q^{(m)}$ of $u_q$ with each
codeword has been formed. Keeping the norm
outside the quantizer, as in norm-explicit quantization \citep{dai2019neq},
puts $\widetilde\ell_i$ in the units of $q^\top k_i/\sqrt d$. The coarse
term $u_q^\top c_{a_i}$ stays in the sum, so all members of a list share its
offset. We use $n_{\mathrm{list}}=512$, $M=8$ and $b=4$, so the index holds
four bytes of codes per token per KV head plus one norm and one list id. We
scan every list at decode time, so every indexed token receives a score. The
inverted lists thus serve both as the coarse quantizer of the residual codes
and as the groups in which the background is aggregated. Probing a subset of
lists, the usual ANN setting, appears only in comparison variants
(Figure~\ref{fig:coverage}); even without a background, scanning every list
selects more accurately (Section~\ref{sec:mechanism}).

\subsection{One score stream, two uses}
\label{sec:method-operator}

Positions are partitioned into a sink prefix $S$, the first tokens, kept
exact as in StreamingLLM \citep{xiao2023streamingllm}; a local window $L$ of
the most recent, not-yet-indexed tokens; and the indexed retrieval region
$R$. The retrieval fraction $\rho$ is the share of $R$ attended exactly;
with $k=\lceil\rho|R|\rceil$,
\begin{equation}
 T(q)=\operatorname{TopK}_{k,\,i\in R}\widetilde\ell_i(q),\qquad
 E=S\cup L\cup T(q),\qquad B=R\setminus T(q).
 \label{eq:partition}
\end{equation}
For $i\in E$ the original $k_i,v_i$ are gathered and the exact logit
$\ell_i=q^\top k_i/\sqrt d$ replaces the approximate one; no second
selection is made. For $i\in B$ the approximate logit is kept. With a common
shift $m_0$ (the maximum over exact and approximate logits),
$e_i=\exp(\ell_i-m_0)$ on $E$ and $\widetilde e_i=\exp(\widetilde\ell_i-m_0)$
on $B$. The background is aggregated per inverted list
$R_j=\{i\in R: a_i=j\}$:
\begin{equation}
 M_j=\sum_{i\in B\cap R_j}\widetilde e_i,\qquad
 \bar v_j=\frac{1}{|R_j|}\sum_{i\in R_j}v_i,\qquad
 \widehat Z_B=\sum_j M_j,\qquad \widehat N_B=\sum_j M_j\,\bar v_j,
 \label{eq:background}
\end{equation}
where an empty list contributes zero. The output is
\begin{equation}
 \widehat y=\frac{N_E+\widehat N_B}{Z_E+\widehat Z_B}.
 \label{eq:output}
\end{equation}
At $\rho=1$, $B$ is empty and Equation~\ref{eq:output} is dense attention.
List means are maintained over all members of $R_j$ at index updates, so the
value side reads $n_{\mathrm{list}}$ list means and none of the $|B|$
unselected rows. Weighting each mean by its members' approximate mass would
make it query-dependent and require reading every unselected value at each
step. The list means also count the selected members, about a fraction
$\rho$ of each list; excluding them changes macro accuracy by $+0.001$
[$-0.010$, $+0.010$] (Appendix~\ref{app:mechanism}).

\subsection{Execution path}
\label{sec:method-exec}

After a dense prefill, the index of each KV head is built from its
prefilled keys by $k$-means, once per prompt. Training the $k$-means problems
of all heads of a layer together \citep{flashkmeans2025} builds the index of
Llama-3.1-8B at 128K in 0.84\,s on one H20, against 6.5\,s when each head is
built separately (Appendix~\ref{app:build}). New tokens stay exact in $L$ and
are encoded into the existing lists every 256 steps, without re-clustering;
each update also refreshes the list means.
Each decode step runs, per layer, the lookup-table construction, the code
scan with a fused per-list reduction, a top-$k$ over the score rows, a gather
of the selected rows, and the combination of
Equation~\ref{eq:output} (Figure~\ref{fig:framework}), replayed as one CUDA
graph \citep{nvidia2025cudagraphs}. Under GQA every query head of a group
shares the codes of its KV head and forms its own lookup table, so one pass
over the codes yields one score row per query head.
Inside vLLM \citep{kwon2023vllm}, whose PagedAttention stores
the KV cache in fixed-size pages, the gather reads the engine's pages in place.

\section{Experiments}
\label{sec:experiments}
\subsection{Setup}
\label{sec:setup}

\paragraph{Models and hardware.}
Llama-3.1-8B-Instruct is the primary model \citep{dubey2024llama3}.
Qwen2.5-14B-1M and Qwen3-30B-A3B extend the speed study to larger models
(Table~\ref{tab:tp}), and Qwen3-30B-A3B also serves as the second model for
quality (Table~\ref{tab:quality}(c)). Gradient-1048k, a Llama-3-8B
checkpoint extended to a 1M-token window, supplies the 256K--512K points
beyond Llama-3.1's 128K window
\citep{qwen2025qwen25_1m,qwen2025qwen3,gradientai2024llama1048k}.
Speed is measured on one NVIDIA H20 GPU (96\,GB) inside the vLLM serving
engine \citep{kwon2023vllm}, and quality in a HuggingFace implementation of
the same operator (Appendix~\ref{app:protocol}).

\paragraph{Benchmarks.}
RULER \citep{hsieh2024ruler} generates synthetic long-context tasks at a
chosen length; at 128K we use four core tasks, core-4 (single- and
multi-key needle retrieval, question answering, variable tracking), and four
extended tasks, extended-4 or ext-4 (multi-query and multi-value needles, common- and
frequent-word extraction). InfiniteBench \citep{zhang2024infinitebench} contributes six tasks at 128K,
including question answering and multiple choice over full-length novels;
LongBench \citep{bai2024longbench} contains sixteen English tasks on real
documents, run at native lengths for the comparison with PQCache and
MagicPIG. AIME-24 and MATH-500 test long generation on
DeepSeek-R1-Distill-Llama-8B. Passkey retrieval, which asks for a random
number hidden in the context \citep{kamradt2023needle}, is the correctness check attached to speed measurements. Scores are macro accuracy, the mean of
per-task accuracies (Appendix~\ref{app:protocol}).

\paragraph{Budgets and baselines.}
The retrieval fraction $\rho$ is 1\% unless stated; Table~\ref{tab:quality}(b, c)
uses 2\%, and the mechanism study reports both. Every budgeted baseline attends
exactly as many tokens as PQ-HSA (about 1.5K at 1\% and 2.8K at 2\% for a
128K input). Baselines are full attention (dense attention, timed with the
FlashAttention-3 kernel), truncation (top-$k$ from a subset of probed
IVF lists, unselected tokens at zero weight), SnapKV, sink-and-window,
Quest, RetrievalAttention, PQCache, MagicPIG, ParisKV and SparQ; ParisKV,
RetrievalAttention and SparQ are our reimplementations of the published
procedures in the same evaluation pipeline. Appendix~\ref{app:protocol} lists
the baseline settings.

\paragraph{Statistics and timing.}
All methods are evaluated on the same prompts, and paired differences come
with 95\% bootstrap intervals from 10{,}000 resamples of the prompts within
each task. Speed is reported in two ways: the attention segment, the GPU
time of the decode attention call summed over layers, and the time per
output token (TPOT) of the whole decode step. Table~\ref{tab:speed-protocol} gives the setting of every speed number.

\subsection{Quality against full attention and budgeted baselines}
\label{sec:quality}
\begin{table}[t]
\centering
\caption{Quality at 128K, higher is better. (a) Llama-3.1-8B-Instruct, macro
accuracy at a 1\% budget; bold marks the best budgeted method. (b, c) All
methods on the same RULER prompts at a 2\% budget, on Llama-3.1-8B (b) and
Qwen3-30B-A3B (c); $\Delta$ = PQ-HSA $-$ method with its paired 95\%
interval, bold where the interval excludes zero. $\dagger$ full attention;
reprod.\ our reimplementation.}
\label{tab:quality}
\footnotesize
\setlength{\tabcolsep}{5pt}
\begin{tabular}{@{}lccc@{}}
\toprule
\multicolumn{4}{@{}l}{(a) 1\% budget} \\
Method & RULER core-4 & RULER ext-4 & InfiniteBench \\
\midrule
full attention$^{\dagger}$ & 0.985 & 0.784 & 0.655 \\
PQ-HSA & \textbf{0.950} & \textbf{0.777} & \textbf{0.646} \\
Quest & 0.915 & 0.641 & 0.575 \\
SnapKV & 0.885 & 0.621 & 0.512 \\
sink + window & 0.050 & 0.207 & 0.105 \\
\bottomrule
\end{tabular}

\vspace{6pt}
\begin{tabular}{@{}lcc@{}}
\toprule
\multicolumn{3}{@{}l}{(b) Llama-3.1-8B, 2\% budget, RULER ext-4} \\
Method & macro & $\Delta$ [95\% CI] \\
\midrule
PQ-HSA & 0.795 & --- \\
full attention$^{\dagger}$ & 0.779 & $+0.016$ [$-0.011$, $+0.044$] \\
ParisKV (reprod.) & 0.738 & $\mathbf{+0.056}$ [$+0.022$, $+0.092$] \\
SnapKV & 0.646 & $\mathbf{+0.148}$ [$+0.097$, $+0.200$] \\
\midrule
\multicolumn{3}{@{}l}{(c) Qwen3-30B-A3B, 2\% budget, ext-4 + single needle} \\
\midrule
PQ-HSA & 0.905 & --- \\
full attention$^{\dagger}$ & 0.914 & $-0.009$ [$-0.021$, $+0.003$] \\
SnapKV & 0.809 & $\mathbf{+0.096}$ [$+0.071$, $+0.120$] \\
Quest & 0.753 & $\mathbf{+0.151}$ [$+0.117$, $+0.189$] \\
\bottomrule
\end{tabular}
\end{table}

Table~\ref{tab:quality}(a) reports macro accuracy at a fixed 1\% budget, on
RULER's synthetic tasks and on InfiniteBench's natural and synthetic ones.
PQ-HSA is the most accurate budgeted method on all three Llama-3.1-8B
benchmark groups and is close to full attention on RULER extended-4 and
InfiniteBench. On paired extended-4 prompts it leads Quest by 0.136
[0.105, 0.165] and SnapKV by 0.156 [0.121, 0.190], and differs from full attention by
$-0.007$ [$-0.031$, $+0.017$]. On RULER core-4 it scores 0.950 (0.970 at
2\%), above our RetrievalAttention reimplementation (0.925;
Appendix~\ref{app:design}) and IVF-probed truncation (0.645).
In long chain-of-thought generation (AIME-24 and MATH-500 with
DeepSeek-R1-Distill-Llama-8B), the prompt is short and the context is
produced during decoding, so the index is extended many times per problem;
there PQ-HSA scores 0.562 and full attention 0.554.

Table~\ref{tab:quality}(b, c) evaluates every method on the same 128K
prompts at a matched 2\% budget. On both models PQ-HSA is more accurate than
every budgeted baseline, with intervals that exclude zero, and on both the
interval against full attention contains zero. PQ-HSA also leads our
SparQ reimplementation \citep{ribar2024sparq} by 0.049 [0.014, 0.086] at a
1\% budget (Appendix~\ref{app:design}).
On the sixteen LongBench tasks at their native lengths (0--100 scale),
PQ-HSA scores 49.1, full attention 48.8, PQCache 47.6 and MagicPIG 48.6
(Appendix~\ref{app:design}).

\subsection{Where the background term's accuracy comes from}
\label{sec:mechanism}
\begin{table}[t]
\centering\small
\caption{Mechanism variants at 128K, in the notation of
Section~\ref{sec:method-operator}; $\bar v_{\mathrm{global}}$ is the
token-weighted global value mean, a logit of $-\infty$ drops the background
(truncation), and --- marks no background term. $^\dagger$128 probed lists (Appendix~\ref{app:mechanism}).}
\label{tab:mechanism-arms}
\begin{tabular}{@{}llll@{}}
\toprule
Variant & Selector & Background logits & Background values \\
\midrule
A & top-$k$, probed lists$^\dagger$ & $-\infty$ & --- \\
T & top-$k$, all lists & $-\infty$ & --- \\
\textbf{B (PQ-HSA)} & top-$k$, all lists & $\widetilde\ell_i$ & $\bar v_j$ \\
C & top-$k$, all lists & $\widetilde\ell_i$ & $\bar v_{\mathrm{global}}$ \\
C2 & top-$k$, all lists & $\operatorname{mean}_{i\in T}\widetilde\ell_i$ & $\bar v_{\mathrm{global}}$ \\
C2b & top-$k$, all lists & $\min_{i\in T}\ell_i$ & $\bar v_{\mathrm{global}}$ \\
Dense & all tokens & --- & --- \\
\bottomrule
\end{tabular}
\end{table}

To see which part of the operator carries the accuracy, we change one
component at a time (Table~\ref{tab:mechanism-arms}), on five 128K RULER
tasks (extended-4 and question answering). B is PQ-HSA. T keeps
B's selector and budget and drops the background, so B versus T measures the
background term itself. C replaces the list value means by the single global
mean value that SparQ uses \citep{ribar2024sparq}, and C2b further gives
every background token one shared logit, the smallest exact logit of the
selected set, so C versus C2b measures what the PQ logits add. A is truncation with the usual IVF
probing of a few lists, so T versus A measures the gain from scoring every
token.

\begin{figure}[t]
\centering
\includegraphics[width=\linewidth]{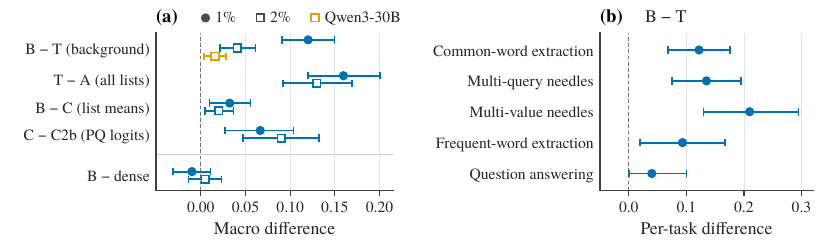}
\caption{Paired differences between the variants of
Table~\ref{tab:mechanism-arms} at 128K, 95\% bootstrap intervals, zero
marked. (a) Macro differences for five contrasts on Llama-3.1-8B at 1\% and
2\% retrieval; the B minus T row adds Qwen3-30B at 2\%. (b) Per-task
B minus T on Llama-3.1-8B at 1\%. Per-task scores are in
Appendix~\ref{app:mechanism}.}
\label{fig:mechanism}
\end{figure}

Removing the background under the same selector lowers accuracy on both
models. On Llama-3.1-8B, B minus T is $+0.120$ [0.091, 0.150] at 1\% and
$+0.041$ [0.021, 0.061] at 2\%, and at 1\% B is ahead of T on every task (Figure~\ref{fig:mechanism}(b)),
including multi-value needles ($+0.210$). On Qwen3-30B at 2\%, B minus T is
$+0.016$ [0.003, 0.029]. Scoring every token also lifts truncation itself,
by 0.160 [0.120, 0.200] (T minus A).
On Llama-3.1-8B both components of the background term contribute, with
intervals that exclude zero (Figure~\ref{fig:mechanism}(a)): list value means
add 0.02--0.03 in macro accuracy over a global mean (B minus C), and
PQ logits add 0.067 at 1\% and 0.090 at 2\% over one shared logit
(C minus C2b), up to 0.436 on common-word extraction at 1\%. With both
in place, B stays close to full attention: B minus full attention is $-0.010$ [$-0.031$, $+0.011$] at 1\% and $+0.005$ [$-0.014$, $+0.023$] at 2\%. Measured directly, the background halves the attention-output
error of truncation: the median relative error to full attention is 0.063
for B against 0.124 for T, and list value means keep it below that of a
global mean (0.070 for C; Appendix~\ref{app:mechanism}).

\subsection{Speed, context length, and the empirical cost model}
\label{sec:speed}

\paragraph{Context length.}
Figure~\ref{fig:headline}(b) reports the attention-segment speedup over the
FlashAttention-3 decode kernel on one H20, and the speedup grows with context
length. Near 128K, Llama-3.1-8B reaches $1.58\times$ when the plugin reads
vLLM's KV pages in place (Section~\ref{sec:deployment}) and $1.76\times$ when
PQ-HSA gathers from its own copy of the KV cache; Gradient-1048k, with the
same architecture, reaches $2.72\times$ at 512K. At 512K the whole decode
step is faster as well: PQ-HSA generates a token in 31.0\,ms against
54.5\,ms for dense attention (Table~\ref{tab:speed-protocol}), as
attention's share of a dense decode step on these 8B models rises from about
half at 128K to about three quarters at 512K. On 512K multi-query needle
retrieval, PQ-HSA gives the same score as full attention on every prompt
(0.79; Appendix~\ref{app:protocol}). The advantage holds with concurrent requests
and against dense attention over an FP8 KV cache ($1.37\times$ at 128K;
Table~\ref{tab:batch-fp8}).

\begin{figure}[t]
\centering
\includegraphics[width=\linewidth]{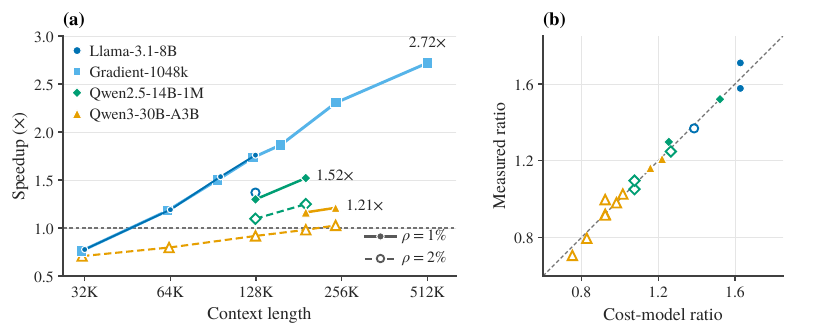}
\caption{Speedup across model sizes and the empirical cost model. (a)
Attention-segment speedup (dense/PQ-HSA) on one H20 against context length at
budgets of 1\% (solid, filled markers) and 2\% (dashed, hollow markers); the Llama-3.1-8B and Gradient-1048k lines at 1\% extend those of
Figure~\ref{fig:headline}(b) to 32K. (b) Measured ratio against the fitted cost model
for the 17 measurements of Table~\ref{tab:tp}; repeated measurements show the median.}
\label{fig:law}
\end{figure}

\paragraph{Model size.}
Figure~\ref{fig:law}(a) adds Qwen2.5-14B-1M, with 8 KV heads and $G=5$, and
Qwen3-30B-A3B, with 4 KV heads and $G=8$. On every model the ratio grows with
context, and each model is faster than dense attention above its break-even
length $L^{*}$ (Equation~\ref{eq:breakeven}). At 1\%, Qwen2.5-14B reaches
$1.52\times$ at 192K, and Qwen3-30B reaches $1.16\times$ at 192K and
$1.21\times$ at 250K. Fewer KV heads and larger groups move $L^{*}$ to longer
contexts.

\paragraph{Cost model.}
For a model with $H_{\mathrm{kv}}$ KV heads, GQA group size $G$ and context
length $L$, we fit the per-layer times
$t_{\mathrm{dense}}=d_0+aH_{\mathrm{kv}}L$ and
$t_{\mathrm{pq}}=F+cH_{\mathrm{kv}}GL$.
The dense kernel streams $H_{\mathrm{kv}}$ heads of $L$ tokens once, so its
time grows with $H_{\mathrm{kv}}L$ above a fixed launch cost $d_0$. PQ-HSA
forms one lookup table and one score row per query head, $H_{\mathrm{kv}}G$
rows per layer, so its time grows with $H_{\mathrm{kv}}GL$ above a fixed
per-layer overhead $F$, and $c$ is the per-row, per-token cost of the scan and
of attending the selected rows. We fit $F$ and $d_0$ per model and $c$ per
budget. Fitted to the 17 measurements of
Table~\ref{tab:tp}, the model stays within 7.4\% of every measured ratio
(Figure~\ref{fig:law}(b); constants in Appendix~\ref{app:law}). Setting the two
times equal gives
\begin{equation}
 L^{*}=\frac{F-d_0}{H_{\mathrm{kv}}(a-cG)},
 \label{eq:breakeven}
\end{equation}
with a crossover when $F>d_0$ and $a>cG$. Above $L^{*}$ the ratio rises toward
$a/(cG)$, so the group size bounds the attainable speedup and the number of KV
heads sets how early the crossover comes. At 1\%, the fitted break-even
lengths are 57K for Llama-3.1-8B, 85K for Qwen2.5-14B and 100K for Qwen3-30B
(both budgets in Appendix~\ref{app:law}).

\subsection{Serving-stack integration}
\label{sec:deployment}

The plugin replaces vLLM's single-token decode attention call without
changing the engine source. Prefill and multi-token steps keep the engine's
own attention, so PQ-HSA acts on the memory-bound decode phase and runs
alongside concurrent requests and prefix caching, which reuses the KV cache of
a shared prompt prefix; the gather reads the engine's paged KV cache in place.
The same package runs on vLLM 0.8.5 and 0.29, which differ in their
attention-backend interface and KV layout, and gathers bit-identical rows on
both (Appendix~\ref{app:systems}). The plugin speeds up the attention call by
$1.58\times$ on Llama-3.1-8B at 128K on vLLM 0.8.5; on vLLM 0.29 the speedup
again grows with context length, to $1.89\times$ at 256K and $2.22\times$ at
512K on Gradient-1048k (Table~\ref{tab:speed-protocol}).

\section{Limitations}
\label{sec:limitations}

Our quality evaluation centers on two GQA models with a 128K
window, and a broader study over longer native contexts and other attention
designs, such as multi-head latent attention, would show how the background
estimate transfers across architectures. Our speed evaluation
targets single-GPU deployment, where the advantage lies in the memory-bound
regime of long contexts; with short contexts, few KV heads or large GQA groups,
a fixed per-layer overhead and the per-query-row scan dominate, and the fitted cost model of Section~\ref{sec:speed} places this boundary
for the measured models.
Reducing that overhead would widen the regime where PQ-HSA is faster.

\section{Conclusion}
\label{sec:conclusion}

PQ-HSA reuses the IVF-PQ scores that rank the cached keys as logits for the
unselected tokens, aggregated per list with list value means under one
softmax with the exact selection. At 128K it is more accurate than Quest and SnapKV at the same budget on Llama-3.1-8B and Qwen3-30B-A3B and approaches full attention on both, and under the same selector the background term adds 0.12 macro accuracy at 1\% retrieval. In the same setting inside vLLM the decode attention call runs $1.6\times$ faster than
the FlashAttention-3 kernel, and its advantage grows with context length; a fitted cost model summarizes when the
speedup appears across 8B, 14B and 30B models, and a vLLM plugin runs the operator on two engine versions without engine
changes. Because the operator asks the index only for a cheap score of every
key, other quantizers and indexes from vector search can take the place of
IVF-PQ.

\subsection*{AI use statement}
We used generative AI tools to help organize and monitor experiments, write
analysis and checking code, draft and revise the manuscript, and prepare
figures and tables. They also produced internal reviews of drafts, which are
separate from conference peer review. All reported measurements and
statistical summaries are backed by recorded execution outputs and analysis
code that the authors checked. We take responsibility for the final content of
this work, including text, claims and artifacts produced with the aid of
generative AI.

\subsection*{Reproducibility statement}
The code at
\url{https://github.com/KunmingSHAO/pqhsa_release} contains the operator and
its CUDA kernels, the vLLM plugin for both engine versions, the quality and
speed evaluation scripts with the configuration behind every speed number, the
paired-bootstrap script and CPU unit tests. Public model
checkpoints and prompt generators are identified in
Appendix~\ref{app:protocol}, which also lists budgets, index settings, clocks
and sample sizes; variant definitions and per-task results are in
Appendix~\ref{app:mechanism}, fitted constants in Appendix~\ref{app:law} and
serving-stack settings in Appendix~\ref{app:systems}. Run scripts record
seeds, engine settings, precision and timing boundaries.

\bibliography{references}
\bibliographystyle{iclr2027_conference}

\clearpage
\appendix
\section{Evaluation protocol}
\label{app:protocol}

\paragraph{Tasks.}
RULER tasks follow the released generators at 131{,}072 tokens with a noise
haystack \citep{hsieh2024ruler}: single needle (\texttt{niah\_s}), multi-key
needle (\texttt{niah\_mk}, four keys), multi-query needle
(\texttt{niah\_multiquery}, four queries), multi-value needle
(\texttt{niah\_multivalue}, four values), common-word extraction
(\texttt{cwe}: ten common words at frequency 30 against three uncommon words
at frequency 3, one few-shot example), frequent-word extraction
(\texttt{fwe}, $\alpha=2$, three answers), variable tracking (\texttt{vt}) and
a single-fact question-answering task (\texttt{qa}, a city--year fact). Output caps depend
on the study: the multi-query and multi-value needle runs of the mechanism
study use 64 tokens; the simpler needle and question-answering settings use
16, frequent-word extraction 64 and common-word extraction 128.
InfiniteBench \citep{zhang2024infinitebench} contributes key--value
retrieval, long-book multiple choice, long-book question answering, math
find, number string and passkey at 128K. LongBench uses the sixteen English
tasks with official metrics and output caps \citep{bai2024longbench}. Long
chain-of-thought generation uses AIME-24 and MATH-500 with greedy decoding
and a 16K generation cap on DeepSeek-R1-Distill-Llama-8B; the paired unit is
the problem. Generations average 6.8K tokens, and the index is extended about
26 times per problem. The 512K check on Gradient-1048k runs the RULER generators at 524{,}288
tokens in the vLLM setting of the 512K speed measurement, with six prompts
each of multi-query needle retrieval and common-word extraction. On
multi-query needles, PQ-HSA and dense attention score 0.79 with the same
score on every prompt; common-word extraction is near zero for dense
attention itself at this length (0.02, PQ-HSA 0.03).

\paragraph{Budgets.}
The exact set contains four sink tokens, a 128-token local window and a
retrieved portion controlled by $\rho$. Final budgets are rounded to whole
pages of 16 tokens: 1{,}456 tokens at 1\% and 2{,}752 at 2\% for the 128K
setting. Budgets are token counts in the exact set. Because $\rho$ is a
fraction of the indexed region, the exact set grows with the context, to
about 5.4K tokens at 512K and 1\%. SnapKV
receives the same number of tokens (observation window 32, pooling kernel 7,
max pooling); Quest receives the same number of tokens in pages of 16;
sink-and-window keeps the four sink tokens and the most recent tokens up to
the budget. In the mechanism study (Appendix~\ref{app:mechanism}) the
truncation variant A selects from IVF-probed candidates (128 lists), whereas B, C, C2 and C2b rank every indexed
token by its PQ score.
SparQ follows its Algorithm~1 \citep{ribar2024sparq}: it scores tokens from
the $r=32$ largest query components (summed over the GQA group), attends the
same number of tokens as PQ-HSA exactly, and gives the remaining mass to the
running mean of all values; the variant without this mean, which the SparQ
authors use for GQA models, scores lower on the same prompts (0.751 against
0.777). ParisKV uses 16 subspaces, collision fraction $0.30$ and $\beta=0.08$ for its
first stage and reranks to the same final token budget; the tokens it selects
need not coincide with PQ-HSA's. This reimplements the published ranking
procedure. The
LongBench comparison with PQCache and MagicPIG
(Table~\ref{tab:retrieval-family}) uses a separate BF16 pipeline at native
task lengths with per-prompt truncation caps; its 800 prompts reach about
65.4K tokens. There PQCache runs at compression ratio 0.20 and MagicPIG with $K=11$, $L=300$.

\paragraph{Implementation and serving stack.}
Measurements use CUDA 12.4. The quality runs use an FP16 HuggingFace implementation in which PQ-HSA
replaces the decode attention of every layer after a dense prefill; prefill
is chunked at 8{,}192 tokens. The index uses 512 lists, eight subspaces of
four bits, residual encoding, one coarse and two PQ training iterations
(four and eight on Qwen3-30B-A3B),
direction normalization and a 1{,}024-token block size for top-$k$; new
tokens are encoded every 256 steps. Speed runs use vLLM~0.8.5.post1 with the
FlashAttention-3 backend, FP16 weights and KV, and the operator
installed through the plugin of Section~\ref{sec:deployment}; the
vLLM~0.29.0 runs use the same plugin with FP16 KV and the engine's default
FlashAttention backend. The attention-segment and decode wall-clock runs use
one GPU.

\paragraph{Statistics.}
Every comparison between two methods is computed on the same prompts. For a
paired difference we resample prompts within each task with replacement
10{,}000 times (seed 20260828), recompute the macro over tasks, and report
the 2.5th and 97.5th percentiles. Sample sizes are 50 prompts per task for
the primary Llama-3.1-8B RULER studies; the mechanism study lists its
per-variant counts in Appendix~\ref{app:mechanism}. Other studies use 25 prompts per task for the comparisons of
Table~\ref{tab:quality}(b, c), 200 prompts for the RULER-core comparisons at 1\%, and 130 problems for the
chain-of-thought study. 

\paragraph{Clocks.}
The attention segment is measured with CUDA events around the decode
attention call on the compute stream and summed over layers; for PQ-HSA it
covers all of its decode kernels: the lookup-table construction, scan,
top-$k$, gather and combination. Time per output token (TPOT) is the host
wall time per generated token (Table~\ref{tab:speed-protocol}).

\section{Mechanism study: variants and per-task results}
\label{app:mechanism}

\begin{figure}[t]
\centering
\includegraphics[width=\linewidth]{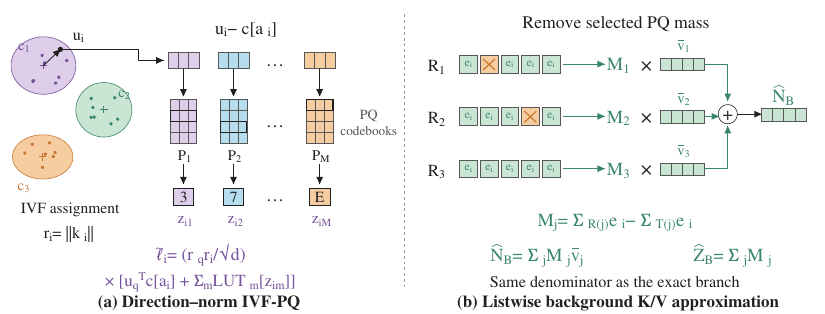}
\caption{(a) Direction--norm residual IVF-PQ: a key direction is assigned to
an IVF centroid and its residual is encoded in $M$ PQ subspaces;
multiplying by the key norm restores logit units. (b) Listwise background
aggregation: each list contributes the approximate mass of its unselected
tokens, weighted by the list's value mean. Here $e_i=\exp(\widetilde\ell_i-m_0)$, and
$R(j)$ and $T(j)$ denote the indexed and selected tokens of list $j$. Clusters and tensor sizes are schematic.}
\label{fig:hybrid}
\end{figure}

Figure~\ref{fig:hybrid} details the index and the listwise aggregation of
the background term.

\paragraph{Variant definitions and execution checks.}
Table~\ref{tab:mechanism-arms} defines the variants: A (truncation),
B (PQ-HSA), C (global value mean) and the shared-logit variants C2 and C2b.
Bx keeps B's selector, background logits and normalizer and replaces each
list mean by the mean of the list's unselected members,
$\bar v_j^{B}=(|R_j|\,\bar v_j-\sum_{i\in T\cap R_j}v_i)/|B\cap R_j|$,
formed per query head from the rows already gathered for the exact set; on
Llama-3.1-8B at 1\%, Bx minus B is $+0.001$ [$-0.010$, $+0.010$] over 250
paired prompts.
C uses a token-count-weighted global value mean accumulated in FP32; C2 and
C2b keep this mean and give all background tokens one shared logit, set
after selection as in Table~\ref{tab:mechanism-arms}; A uses the IVF-probed selector with zero
background. T, B, Bx, C, C2 and C2b use the same selector and budget and differ only in how the
background term is formed. We verified that each variant's code path
is executed.

\paragraph{Attention-output error.}
On two 128K prompts (common-word extraction and multi-query needles) we
recompute, for every layer, query head and the first eight decode steps of
Llama-3.1-8B at 1\%, the relative error $\|\widehat y-y\|/\|y\|$ of each
variant's attention output against full attention, 16{,}384 rows in total.
The median (90th percentile) is 0.124 (0.476) for T, 0.070 (0.281) for C and
0.063 (0.228) for B. B has a lower error than T in 99.1\% of the rows and
than C in 84.0\%, in both halves of the network.

\paragraph{Per-task results.}
Tables~\ref{tab:mech-8b} and~\ref{tab:mech-30b} report every variant on the
same prompts, 25 per task, so rows can be compared directly; the paired
contrasts in the main text use all the prompts each pair shares.

\begin{table}[ht]
\centering\footnotesize
\setlength{\tabcolsep}{4pt}
\caption{Per-variant quality on Llama-3.1-8B at 128K (higher is better). A
selects from IVF-probed candidates; T, B, Bx, C, C2 and C2b rank every indexed
token by its PQ score, and T drops the background. MQ/MV: multi-query/multi-value needles; CWE/FWE:
common-/frequent-word extraction; QA: question answering. All rows use the same 25 prompts per task; bold marks
PQ-HSA (B).}
\label{tab:mech-8b}
\begin{tabular}{@{}llrrrrrrr@{}}
\toprule
$\rho$ & Variant & $n$/task & MQ & MV & CWE & FWE & QA & Macro \\
\midrule
1\% & Dense & 25 & 0.970 & 0.960 & 0.332 & 0.853 & 0.960 & 0.815 \\
1\% & B & 25 & 0.930 & 0.920 & 0.492 & 0.827 & 0.960 & \textbf{0.826} \\
1\% & Bx & 25 & 0.930 & 0.920 & 0.500 & 0.827 & 0.960 & 0.827 \\
1\% & T & 25 & 0.780 & 0.720 & 0.360 & 0.760 & 0.920 & 0.708 \\
1\% & C & 25 & 0.880 & 0.880 & 0.472 & 0.720 & 0.920 & 0.774 \\
1\% & C2b & 25 & 0.910 & 0.860 & 0.036 & 0.733 & 1.000 & 0.708 \\
1\% & A & 25 & 0.550 & 0.390 & 0.332 & 0.747 & 0.680 & 0.540 \\
1\% & C2 & 25 & 0.680 & 0.650 & 0.004 & 0.000 & 0.960 & 0.459 \\
\midrule
2\% & Dense & 25 & 0.970 & 0.960 & 0.332 & 0.853 & 0.960 & 0.815 \\
2\% & B & 25 & 0.970 & 0.940 & 0.468 & 0.800 & 0.960 & \textbf{0.828} \\
2\% & T & 25 & 0.910 & 0.920 & 0.364 & 0.813 & 0.920 & 0.785 \\
2\% & C & 25 & 0.950 & 0.930 & 0.420 & 0.787 & 0.920 & 0.801 \\
2\% & C2b & 25 & 0.950 & 0.930 & 0.024 & 0.693 & 0.960 & 0.711 \\
2\% & A & 25 & 0.700 & 0.590 & 0.376 & 0.773 & 0.840 & 0.656 \\
2\% & C2 & 25 & 0.840 & 0.760 & 0.260 & 0.053 & 0.920 & 0.567 \\
\bottomrule
\end{tabular}
\end{table}

\begin{table}[ht]
\centering\footnotesize
\setlength{\tabcolsep}{4pt}
\caption{Per-variant quality on Qwen3-30B-A3B at 128K and 2\% retrieval. All rows use the same 25 prompts per task. Single: single needle; other
columns and bold as in Table~\ref{tab:mech-8b}.}
\label{tab:mech-30b}
\begin{tabular}{@{}lrrrrrrr@{}}
\toprule
Variant & $n$/task & Single & MQ & MV & CWE & FWE & Macro \\
\midrule
Dense & 25 & 1.000 & 1.000 & 1.000 & 0.648 & 0.920 & 0.914 \\
B & 25 & 1.000 & 1.000 & 1.000 & 0.616 & 0.907 & \textbf{0.905} \\
T & 25 & 1.000 & 1.000 & 1.000 & 0.536 & 0.907 & 0.889 \\
\bottomrule
\end{tabular}
\end{table}

\section{Cost model, fitted constants and index build}
\label{app:law}

\paragraph{Measurement protocol.}
Table~\ref{tab:speed-protocol} lists the run behind each speedup in the paper.

\begin{table}[ht]
\centering
\caption{Protocol of every speedup quoted in the paper.
All points use one H20, a single request, a 1\% retrieval budget, fp16 weights
and KV, and vLLM's FlashAttention-3 kernel for
dense attention. CUDA events time the attention call summed over layers, including all of
PQ-HSA's decode kernels. The last row reports time per output token for the whole decode step
(median process). Rows with $3\times64$ steps report the median of three processes.
``Copy'' gathers from a separate copy of the KV cache held by PQ-HSA;
``paged'' reads the engine's KV pages in place.}
\label{tab:speed-protocol}
\footnotesize
\setlength{\tabcolsep}{3pt}
\begin{tabular}{@{}llllrr@{}}
\toprule
Quoted in & Model, input tokens & vLLM & KV & Steps & Ratio \\
\midrule
Tab.~\ref{tab:tp} & Llama-3.1-8B, 130{,}400 & 0.8.5 & copy & $3\times64$ & $1.71\times$ \\
Fig.~\ref{fig:headline}(b) & Llama-3.1-8B, 130{,}400 & 0.8.5 & copy & 32 & $1.76\times$ \\
Figs.~\ref{fig:headline}(b), \ref{fig:law}(a) & Llama / Gradient, 32K--250K & 0.8.5 & copy & 32--64 & 0.76--$2.31\times$ \\
\S\ref{sec:introduction}, \S\ref{sec:deployment}, Tab.~\ref{tab:tp} & Llama-3.1-8B, 130{,}400 & 0.8.5 plugin & paged & $3\times64$ & $\mathbf{1.58\times}$ \\
Fig.~\ref{fig:headline}(b) & Gradient-1048k, 524{,}288 & 0.8.5 & paged & 32 & $\mathbf{2.72\times}$ \\
Fig.~\ref{fig:law}(a) & Qwen2.5-14B, 195{,}936 & 0.8.5 & paged & $3\times64$ & $1.52\times$ \\
Fig.~\ref{fig:law}(a) & Qwen3-30B, 195{,}936 & 0.8.5 & paged & $3\times64$ & $1.16\times$ \\
Fig.~\ref{fig:law}(a) & Qwen3-30B, 249{,}328 & 0.8.5 & paged & $3\times64$ & $1.21\times$ \\
\midrule
\S\ref{sec:deployment} & Llama-3.1-8B, 32K & 0.29 plugin & paged & 256 & $1.07\times$ \\
\S\ref{sec:deployment} & Gradient-1048k, 256K & 0.29 plugin & paged & 256 & $1.89\times$ \\
\S\ref{sec:deployment} & Gradient-1048k, 524{,}288 & 0.29 plugin & paged & 256 & $2.22\times$ \\
\midrule
\S\ref{sec:speed} & Gradient-1048k, 524{,}288 & 0.8.5 & paged & 256/512 & $\mathbf{1.76\times}$ \\
\bottomrule
\end{tabular}
\end{table}

Table~\ref{tab:batch-fp8} extends the single-request protocol to concurrent
requests and to an FP8 KV cache. With an FP8 KV cache (e4m3,
bf16 weights, FlashAttention-3), dense attention reads half the bytes; PQ-HSA keeps fp16 keys and values and
remains faster, by $1.37\times$ at 128K.

\begin{table}[ht]
\centering
\caption{Attention-segment speedup (dense/PQ-HSA) per decode step on the
concurrent-request path (Llama-3.1-8B, 1\% budget, vLLM~0.8.5 with paged KV; median
of three processes, lower median of four for 64K with 4 requests), and against dense
attention over an FP8 KV cache at 128K (median of three processes per arm).}
\label{tab:batch-fp8}
\footnotesize
\setlength{\tabcolsep}{6pt}
\begin{tabular}{@{}llr@{}}
\toprule
Context & Setting & Speedup \\
\midrule
64K & 1 request & $1.03\times$ \\
64K & 2 requests & $\mathbf{1.71\times}$ \\
64K & 4 requests & $1.39\times$ \\
128K & 1 request & $1.53\times$ \\
128K & 2 requests & $\mathbf{2.60\times}$ \\
\addlinespace[2pt]
128K & FP8 KV for dense, 1 request & $1.37\times$ \\
\bottomrule
\end{tabular}
\end{table}

\paragraph{Derivation.}
Per layer, the dense decode kernel streams the $H_{\mathrm{kv}}$ KV heads of
$L$ tokens once, so its time is $t_{\mathrm{dense}}=d_0+aH_{\mathrm{kv}}L$ with
a launch and reduction overhead $d_0$. PQ-HSA reads the packed codes of the
$H_{\mathrm{kv}}$ heads once but forms one lookup table and one score row per
query head, and reduces, ranks and gathers per row, so its time is
$t_{\mathrm{pq}}=F+cH_{\mathrm{kv}}GL$. The fixed per-layer overhead $F$
collects the lookup-table construction, the top-$k$ launch, the per-list
reduction, the graph replay overhead and the part of the per-row work that
does not grow with $L$; $c$ is the per-row per-token cost of the scan and of
attending the $\rho L$ selected rows, and depends on $\rho$. The ratio
$R=t_{\mathrm{dense}}/t_{\mathrm{pq}}$ equals one at $L^{*}$ of
Equation~\ref{eq:breakeven} and tends to $a/(cG)$ as $L\to\infty$.

\begin{table}[t]
\centering
\begin{threeparttable}
\caption{Attention-segment speedup $R$ (dense/PQ-HSA) on one H20 across model
sizes, context lengths $L$ and budgets $\rho$. $H_{\mathrm{kv}}$ is the number
of KV heads and $G$ the GQA group size. KV: ``copy'' gathers from a separate
copy of the KV cache and ``paged'' reads the engine's pages in place
(Table~\ref{tab:speed-protocol}). $R_{\mathrm{model}}$ is the cost model of
Section~\ref{sec:speed} fitted to all rows; bold marks each model's largest measured ratio. Rows at 192K
and 250K and both Llama-3.1-8B rows at 1\% report the median of three processes.}
\label{tab:tp}
\footnotesize
\setlength{\tabcolsep}{4.5pt}
\begin{tabular}{@{}lrrrrlr@{}}
\toprule
Model & $L$ & $\rho$ & $H_{\mathrm{kv}}$ & $G$ & KV & $R$ / $R_{\mathrm{model}}$ \\
\midrule
Llama-3.1-8B & 128K & 1\% & 8 & 4 & copy  & \textbf{1.71} / 1.63 \\
             & 128K & 1\% & 8 & 4 & paged & 1.58 / 1.63 \\
             & 128K & 2\% & 8 & 4 & paged & 1.37 / 1.39 \\
\addlinespace[2pt]
Qwen2.5-14B-1M & 128K & 1\% & 8 & 5 & paged & 1.30 / 1.26 \\
               & 128K & 2\% & 8 & 5 & paged & 1.05 / 1.08 \\
               & 128K & 2\% & 8 & 5 & paged & 1.10 / 1.08 \\
               & 192K & 1\% & 8 & 5 & paged & \textbf{1.52} / 1.52 \\
               & 192K & 2\% & 8 & 5 & paged & 1.25 / 1.27 \\
\addlinespace[2pt]
Qwen3-30B-A3B & 32K  & 2\% & 4 & 8 & paged & 0.71 / 0.75 \\
              & 64K  & 2\% & 4 & 8 & paged & 0.80 / 0.83 \\
              & 128K & 2\% & 4 & 8 & copy  & 1.00 / 0.92 \\
              & 128K & 2\% & 4 & 8 & paged & 0.92 / 0.92 \\
              & 128K & 2\% & 4 & 8 & paged & 0.92 / 0.92 \\
              & 192K & 1\% & 4 & 8 & paged & 1.16 / 1.16 \\
              & 192K & 2\% & 4 & 8 & paged & 0.98 / 0.98 \\
              & 250K & 1\% & 4 & 8 & paged & \textbf{1.21} / 1.22 \\
              & 250K & 2\% & 4 & 8 & paged & 1.03 / 1.02 \\
\bottomrule
\end{tabular}
\end{threeparttable}
\end{table}

\paragraph{Fit.}
The fit uses the 17 single-device measurements of Table~\ref{tab:tp}, two
that gather from a separate KV copy and fifteen that read the engine's pages
in place, with repeated configurations entered as separate points; a point measured in three processes enters with the median of each per-layer time. Each model
has one value of $H_{\mathrm{kv}}G$ on one device, so a length-independent
per-row term is absorbed into $F$. Least squares uses a shared $a$, per-model $d_0$ and $F$, and per-budget $c$,
on the per-layer times; the dense fit has $R^2=0.998$ and the PQ-HSA fit
$R^2=0.985$. The fitted
ratios lie within 7.4\% of the measured ones on every point, with a median
deviation of 1.4\%, and fall on the same side of one as the measurements.

\paragraph{Break-even lengths.}
At $\rho=1\%$ and $2\%$: Llama-3.1-8B, $L^{*}=57$K and 67K; Qwen2.5-14B, 85K
and 109K; Qwen3-30B-A3B, 100K and 217K. The asymptotes $a/(cG)$ are 3.3, 2.7
and 1.7 at $\rho=1\%$ for $G=4$, 5 and 8, and 2.5, 2.0 and 1.2 at $\rho=2\%$.
Every measured point above its model's $L^{*}$ in Table~\ref{tab:tp} and on
the Llama-3.1-8B line of Figure~\ref{fig:headline}(b) is faster than dense
attention.

\paragraph{Index build.}
\label{app:build}
The per-head build of Llama-3.1-8B solves 2{,}304 small $k$-means problems,
one coarse and eight subspace problems for each of its 256 heads, and its
time is dominated by kernel launches and host synchronization. The batched
build solves one coarse problem per layer over its eight heads and one PQ
problem over its 64 head--subspace pairs, computes key norms in FP32 and
re-seeds empty clusters with the rule of the per-head build. On a 128K RULER
prompt it builds all 32 layers in 0.84\,s against 6.53\,s per head on one
H20, and assigns 99.999\% of the keys to the same list and 99.98\% to the
same PQ codes as the per-head build.

\section{Position in the retrieval design space}
\label{app:design}

\begin{table}[ht]
\centering
\caption{Retrieval-family methods compared by design. Index is the state
kept besides the KV cache; ``host'' marks state kept in host memory in the
released stacks. $^{a}$In SparQ's fastest kernel.}
\label{tab:design}
\footnotesize
\setlength{\tabcolsep}{3.5pt}
\begin{tabular}{@{}lllll@{}}
\toprule
Method & Unit & Selection & Unselected tokens & Index \\
\midrule
Quest & page & bound-based top-$k$ & dropped & page bounds \\
ParisKV & token & low-bit pass, rerank & dropped & ANN index \\
PQCache & token & PQ top-$k$ & dropped & PQ codes (host) \\
SparQ & token & $r$ of $d$ key components & one global mean value & second key copy$^{a}$ \\
RetroInfer & cluster & centroid top clusters & centroid logit, value sum; rest dropped & clusters (host KV) \\
Multipole & cluster & centroid top clusters & centroid logit, mean value & key clusters \\
MagicPIG & token & LSH sampling & sampling estimate & LSH tables (host) \\
SnapKV & token & prefill eviction & evicted & none \\
Spec.\ selection & token & 1B draft model & dropped & none \\
\midrule
PQ-HSA & token & IVF-PQ top-$k$ & per-token PQ logit, list mean value & IVF-PQ codes \\
\bottomrule
\end{tabular}
\end{table}

\begin{table}[ht]
\centering
\caption{Quality comparisons with retrieval-family baselines: the sixteen
LongBench tasks at native lengths (up to 65.4K tokens; 0--100 scale, 50
prompts per task) in the BF16 pipeline, and RULER core-4 at 128K and a 1\%
budget in the primary FP16 pipeline, where RetrievalAttention and SparQ
(at 1\%) are our reimplementations under the protocol of
Appendix~\ref{app:protocol}.}
\label{tab:retrieval-family}
\footnotesize
\begin{tabular}{@{}llccc@{}}
\toprule
Method & benchmark & $n$ & method & PQ-HSA \\
\midrule
full attention & LongBench-16 & $16{\times}50$ & 48.8 & \textbf{49.1} \\
PQCache & LongBench-16 & $16{\times}50$ & 47.6 & \textbf{49.1} \\
MagicPIG & LongBench-16 & $16{\times}50$ & 48.6 & \textbf{49.1} \\
\midrule
RetrievalAttention & RULER core-4 & $4{\times}50$ & 0.925 & \textbf{0.950} \\
SparQ & RULER, 5 tasks & $5{\times}25$ & 0.777 & \textbf{0.826} \\
\bottomrule
\end{tabular}
\end{table}

Table~\ref{tab:design} places the operator among the retrieval-family
methods by the unit of selection, how the exact rows are chosen, what the
unselected tokens receive, and the index state. PQ-HSA attends the retrieved
rows plus its exact sink and local regions; the background costs one
reduction over the score rows the selector already produced and
$n_{\mathrm{list}}$ value means. Page-level selection fetches whole pages
around each hit; two-stage rerankers fetch a candidate set several times
larger than the final $k$ at reduced precision before the exact pass; LSH
sampling fetches a sample whose size is set by the estimator's variance;
speculative selection runs a second model.

Request times of the
released PQCache and MagicPIG stacks are compared in
Appendix~\ref{app:extra}.

Table~\ref{tab:retrieval-family} collects the quality comparisons with
PQCache and MagicPIG on LongBench and with our RetrievalAttention
reimplementation on RULER. Settings of the LongBench methods are listed in Appendix~\ref{app:protocol}.

\section{Serving-stack details}
\label{app:systems}
\label{app:agentic}\label{app:mtp}\label{app:mla}

\paragraph{Plugin.}
The package registers itself through vLLM's plugin entry point and modifies
no engine source. When enabled
it wraps the decode attention implementation of the FlashAttention backend,
keeps a per-layer adapter that owns the index, and routes single-token decode
steps through Equation~\ref{eq:output}; prefill chunks and multi-token steps call the original implementation. The adapter gathers
selected rows directly from the engine's paged KV cache, so no second copy of
keys or values is kept. Between vLLM~0.8.5.post1 and 0.29.0 the attention
backend's interface and the paged KV layout changed (keys and values are now
interleaved in a non-contiguous $[\text{blocks}, \text{heads}, \text{block},
2d]$ view); the plugin handles each difference behind a version check, reads
the new layout in place through its strides, and a 48-case test confirms
that the gather is bit-identical to one from the earlier layout and from a
flat buffer.

\section{Additional measurements}
\label{app:extra}

\paragraph{Offload-based baselines.}
End-to-end request time was also compared with the released PQCache and
MagicPIG stacks, which keep part of their state on the host, in BF16 at
126{,}720 input and 4{,}096 output tokens. The paired geometric-mean ratios
PQ-HSA/PQCache and PQ-HSA/MagicPIG are 0.0264 and 0.2448 (about $38\times$ and $4.1\times$ less time). This comparison uses end-to-end
request time in BF16, a different clock from the attention-segment ratios
elsewhere in the paper.

\end{document}